\documentclass{article}

\usepackage[nodisplayskipstretch]{setspace}

\usepackage[final]{nips_2018}

\usepackage[utf8]{inputenc} 
\usepackage[T1]{fontenc}    
\usepackage{url}            
\usepackage{booktabs}       
\usepackage{amsfonts}       
\usepackage{nicefrac}       
\usepackage{microtype}      
\usepackage{graphicx}
\usepackage{multirow}
\usepackage{subcaption}
\usepackage{tablefootnote}
\usepackage[titletoc]{appendix}
\usepackage{natbib}
\usepackage{hyperref}       

\newcommand{\splitcell}[2][c]{\begin{tabular}[#1]{@{}c@{}}#2\end{tabular}}

\title{Resource-Efficient Neural Architect}

\author{
  Yanqi Zhou \\
  \texttt{zhouyanqi@baidu.com} \\
  \And
  Siavash Ebrahimi \\
 \texttt{ebrahimisiavash@baidu.com}
  \And
  Sercan \"{O}. Ar{\i}k\\
  \texttt{sercanarik@baidu.com} \\
  \And
  Haonan Yu\\
  \texttt{haonanyu@baidu.com} \\
  \And
  Hairong Liu\\
  \texttt{liuhairong@baidu.com} \\
  \And
  Greg Diamos\\
  \texttt{gregdiamos@baidu.com} \\
}

\begin{document}

\maketitle

\begin{abstract}
Neural Architecture Search (NAS) is a laborious process. Prior work on automated NAS targets mainly on improving accuracy, but lacks consideration of computational resource use. We propose the Resource-Efficient Neural Architect (RENA), an efficient resource-constrained NAS using reinforcement learning with network embedding. RENA uses a policy network to process the network embeddings to generate new configurations. We demonstrate RENA on image recognition and keyword spotting (KWS) problems. RENA can find novel architectures that achieve high performance even with tight resource constraints. For CIFAR10, it achieves 2.95\% test error when compute intensity is greater than 100 FLOPs/byte, and 3.87\% test error when model size is less than 3M parameters. For Google Speech Commands Dataset, RENA achieves the state-of-the-art accuracy without resource constraints, and it outperforms the optimized architectures with tight resource constraints. 

\end{abstract}

\section{Introduction}

Deep neural networks have demonstrated excellent performance on challenging research benchmarks, while pushing the frontiers of numerous impactful applications such as language translation \citep{MT1}, speech recognition \citep{DS2}, speech synthesis \citep{parallelwavenet}, image recognition \citep{imagerecognition1} and image synthesis \citep{imagegeneration1}. Despite all these advancements, designing neural networks still remains to be a laborious task, requiring extensive experience and expertise. With the motivation of automating the neural network development process while achieving competitive performance, neural architecture search (NAS) has been proposed ~\citep{Real2017,Zoph2016,Liu2017b,Baker2016,Negrinho2017}. For some competitive benchmarks like image recognition, NAS has already yielded very promising results compared to manually designed models. 

Historical trend in artificial intelligence research has been improving the performance of a model on a certain task, without considering resource use metrics, such as model memory, complexity, and power consumption. Larger and deeper neural networks with specially-designed architectures have been developed along this trend. On the other hand, as deep neural networks are starting to be deployed in different applications more widely, resource constraints become crucial besides performance. Specifically, resource-constrained neural network development are motivated by two recent trends: \\
(i) There is a growing interest in optimizing the performance of modern processors for deep neural networks, e.g. \citep{TPU_datacenter}. These specialized architectures typically yield their peak performance for algorithms with high compute intensity. Yet, it has been an uncommon research practice to develop neural network architectures that would yield high compute intensities.\\
(ii) Besides conventional computing platforms like datacenters or personal computers, deep neural networks are being deployed on a wide variety of hardware platforms, such as smartphones, drones, autonomous vehicles, and smart speakers, etc. Such platforms may vary hugely in terms of their computation capabilities, memory capacities or power budgets, as well as the performance targets. Thus, a neural network needs to be re-optimized for every hardware platform it will be deployed on.

Resource constraints exacerbate the challenges of neural network model development, and it is strongly desired to automate this process along the two trends mentioned above. In this paper, we propose a resource-constrained NAS framework, named Resource-Efficient Neural Architect (RENA). Our goal is to automate the process of finding high-performance neural network architectures under different resource constraints with a reasonable amount of search. Our major contributions include:

\begin{enumerate}
    \item We design a policy network with network embedding to adapt existing models progressively, rather than building from scratch.
    \item We introduce a framework for modifying the reward function to tailor target models to meet hardware constraints and propose three simple metrics (model size, compute complexity, and compute intensity) that are interpretable to hardware designers and can be used to guide the search.  
    \item We demonstrate competitive performance for two tasks: (i) image recognition, and (ii) keyword spotting (KWS), even with tight resource constraints.
    
\end{enumerate}

\section{Related Work}

\subsection{Neural Architecture Search}
Automatic neural architecture search (NAS) has been a long-standing research area. Evolutionary algorithms~\citep{Liu.2017.HR,Esteban.2017.LSE,Esteban.2018.REIC,HRforEAS}, are one of the earliest methods used for automatic NAS. NAS has also been studied in the context of Bayesian optimization~\citep{Bergstra.2012.RSH,KK.2018.NASBOO}. Recently, reinforcement learning~\citep{Baker2016,Zoph2016,Baker} has emerged as an effective method for automatic NAS. However, conventional NAS is computationally expensive and time consuming - many results are obtained with a vast amount of computational resources. This renders NAS less realistic for widespread use in research. To address this, efficient automatic NAS with parameter sharing~\citep{Pham2018}, regularized search~\citep{Real2018}, and network morphism~\citep{Elsken.2017.MOAS} are becoming a critical research area. Parameter sharing forces all child models to share weights to eschew training each child model from scratch to convergence. 


\subsection{Resource-Constrained Neural Networks}

Most of the effective approaches to optimize performance under resource constraints still rely on the creativity of the researchers. Among many, some notable ones include attention mechanisms \citep{showattentell}, depthwise-separable convolutions \citep{depthwisesepconv,MobileNet}, inverted residuals \citep{mobilenetv2}, and structured transforms \citep{structuredtransforms}. Aside from the approaches that optimize the neural network architecture that change the type of the layers, common approaches to reduce redundancy indeed use techniques that do not modify the form of the network architecture. These include sparsity regularization \citep{sparsity1}, connection pruning \citep{pruning1}, and reducing the precision of weights and activations \citep{quantization1}. Lastly, training a smaller (student) network to mimic a larger (teacher) network, commonly known as distillation~\citep{distillation,N2N}, has gained traction. For example, in~\citep{parallelwavenet}, distillation is applied to learning an inverse-autoregressive flow model.  

\section{Modeling Resource Use}

Modeling the hardware performance of an algorithm is undoubtedly a challenging task. Our goal in this paper is not to model the performance in the most precise way, but rather to show that when approximate metrics are considered, RENA can efficiently optimize them. For example, for embedded devices, inference latency and power consumption are two important metrics when deploying neural networks. Yet, accurate modeling of them are very challenging - typical approaches depend on various assumptions about the hardware platforms. Instead, we choose to focus on inference metrics that can be precisely quantified in terms of the fundamental operations, and that can also provide interpretable insights. The three metrics that we consider are:\\
(i) \textbf{Model size}: Model size is quantified by the total amount of memory used by the model parameters. For a given neural network, model size depends on the dimensions of the weight tensors, and the precision of each tensor.\footnote{In this paper, we fix the precision of weights to 4 bytes and focus only on the tensor sizes.} Straightforward approaches to reduce the model size may involve reducing the input resolution (e.g. decreasing the number of frequency channels in spectral representation), removing layers, reducing the number of hidden units (e.g. for recurrent cells) or reducing the number of filters (e.g. for convolutions). For a target performance, reduction of model size encourages architectures with more parameter sharing (e.g. depthwise-separable convolutions with short filter sizes) and repetitive computations (e.g. recurrent layers with long sequence lengths and small number of hidden units).\\
(ii) \textbf{Computational complexity}: Computational complexity is quantified by the total number of floating-point operations (FLOPs) (see Appendix \ref{complexity_modeling} for details). Straightforward approaches to reduce the inference complexity are mostly similar to the approaches to reduce the model size, such as reducing the number of hidden units or the number of filters. In general, reduction of complexity encourages models with minimal redundancy (e.g. by joining concetenated linear operations).\\
(iii) \textbf{Compute intensity}: Compute intensity is defined as the average number of FLOPs per data access (i.e. data transfer between the fast and slow memory)\footnote{Note that our definition differs from \citep{TPU_datacenter} as they model the compute intensity per weight access, ignoring input and output data.}. Compute intensity is a measure of how efficiently an algorithm can re-use data. For modern multi-core architectures like GPUs and TPUs~\citep{TPU_datacenter}, it is an indirect measure of how fast the algorithm can be run. In general, if a neural network reuses data, it requires less memory bandwidth and achieves higher compute intensity. High compute intensity encourages neural networks with more locality and often more parallelism. As a simple example, consider matrix-matrix multiplication of an $m \times n$ matrix and an $n \times p$ matrix. The compute intensity would be proportional to $\frac{mnp}{mn + np} = \frac{1}{1/p + 1/m}$. Increasing it would favor for increases in $p$ and $m$. If there is a constraint on their sum, due to the total model size or overfitting considerations, higher compute intensity would favor for $p$ and $m$ values close to each other. One example of a very high compute intensity neural network layer is multi-dimensional convolution with appropriately large channel sizes. On the other hand, recurrent layers used in typical language or speech processing applications, or some recently-popular techniques like multi-branch networks ~\citep{Christian.2015.Inception}, yield low compute intensity.

\section{Architecture Search with Reinforcement Learning}

In this section, we explain the overall RL framework of RENA and the corresponding search space. The framework consists of a policy network to generate actions that define the neural network architecture. The environment outputs the performance of the trained neural network, as well as its resource use. We use policy gradient with accumulated rewards to train the policy network. 

\subsection{Policy Network}
\begin{figure}[!tp]
\centering
\includegraphics[width=0.6\linewidth]{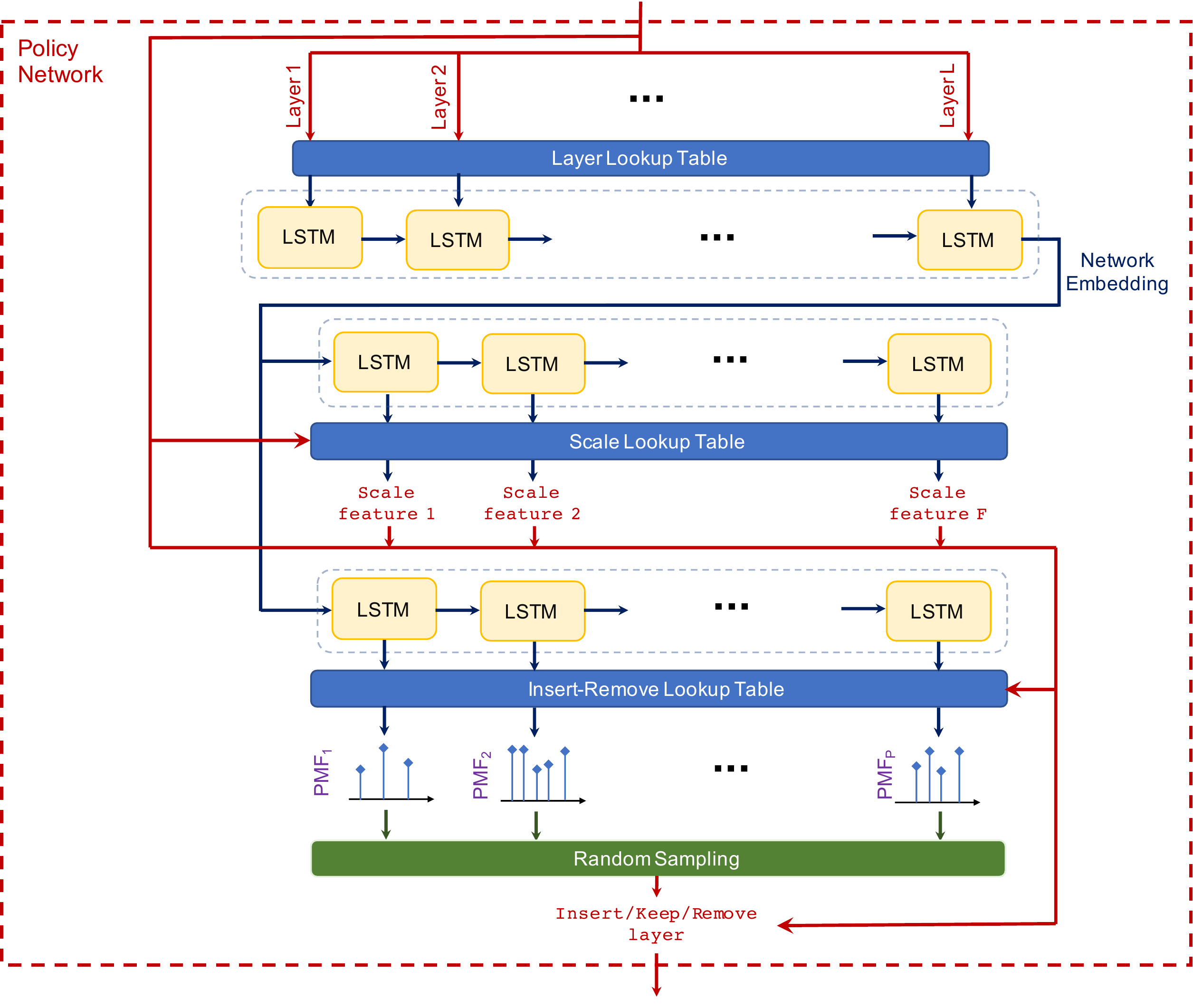}
\caption{Policy Network with Network Embedding: An LSTM-based network transforms an existing neural network configuration into a trainable representation. The trainable representation is fed to a LSTM-based policy network to generate actions.}
\vspace{-1em}
\label{fig:policy-network-with-embedding}
\end{figure}
Policy network, shown in Fig.~\ref{fig:policy-network-with-embedding}, adapts an existing network by modifying its parameters (referred to as the \emph{scale action}), or by inserting a new layer (referred to as the \emph{insert action}), or by removing an existing layer (referred to as the \emph{remove action}). Rather than building the target network from scratch, modifications via these operations allow more sample-efficient search with a simpler architecture. The search can start with any baseline models, a well-designed or even a rudimentary one. 

Policy network uses a \emph{network embedding} to represent the input neural network configuration. Initially, each layer of the input neural network is mapped to layer embeddings by using a trainable lookup table. Then, an LSTM layer (with a state size equal to the number of layers $L$) sequentially processes these layer embeddings and output the network embedding. Next, the network embedding is input to two different LSTMs to define the scale, insert, and remove actions. Scale LSTM outputs the hidden units at every step which correspond to modification of the value of the corresponding feature. Selection from the predefined set of values is done using a lookup table. For example, if the input network consists a convolution layer, one state can change the filter width, the next can change the number of filters and so on. The second LSTM selects between insert, keep, or remove actions based on the output of an additional LSTM state, besides those representing the layer features. Then, either a new layer is inserted and integrated into the scaled neural network or one of the existing layers of the scaled network is removed, or the architecture design remains unchanged. To encourage exploration, the new layers are generated in a stochastic way. Hence, the goal of the insert LSTM is to define the probability mass function (p.m.f.) to sample the features of the new layer to be generated. For each feature, mapping of the LSTM state output to the p.m.f. is done by a lookup table. For example, if there are 3 candidate values for the feature of convolution width, the LSTM state output determines 3 probability values corresponding to them. Finally, the new network is defined by implementing the insert/remove action on top of the scale action.

\subsection{Search Space}

Actions of scale and insert need to be mapped to a search space to define the neural network architectures. We describe the two approaches to define search spaces next.

\begin{figure}[t]
    \begin{subfigure}{0.37\linewidth}
        \centering
        \includegraphics[width=0.7\linewidth]{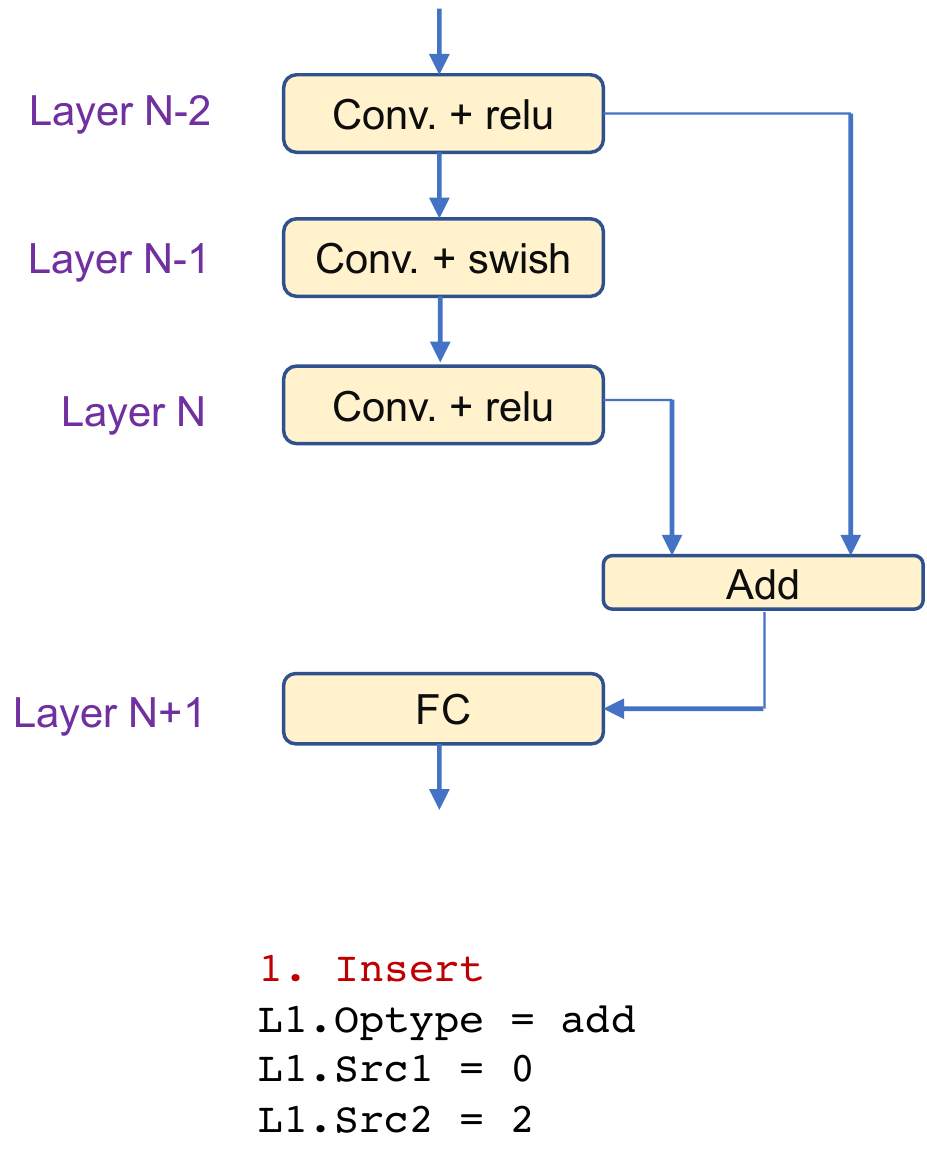}
        \caption{}
        \label{fig:two-ops}
    \end{subfigure}
    \begin{subfigure}{0.6\linewidth}
        \centering
        \includegraphics[width=0.99\linewidth]{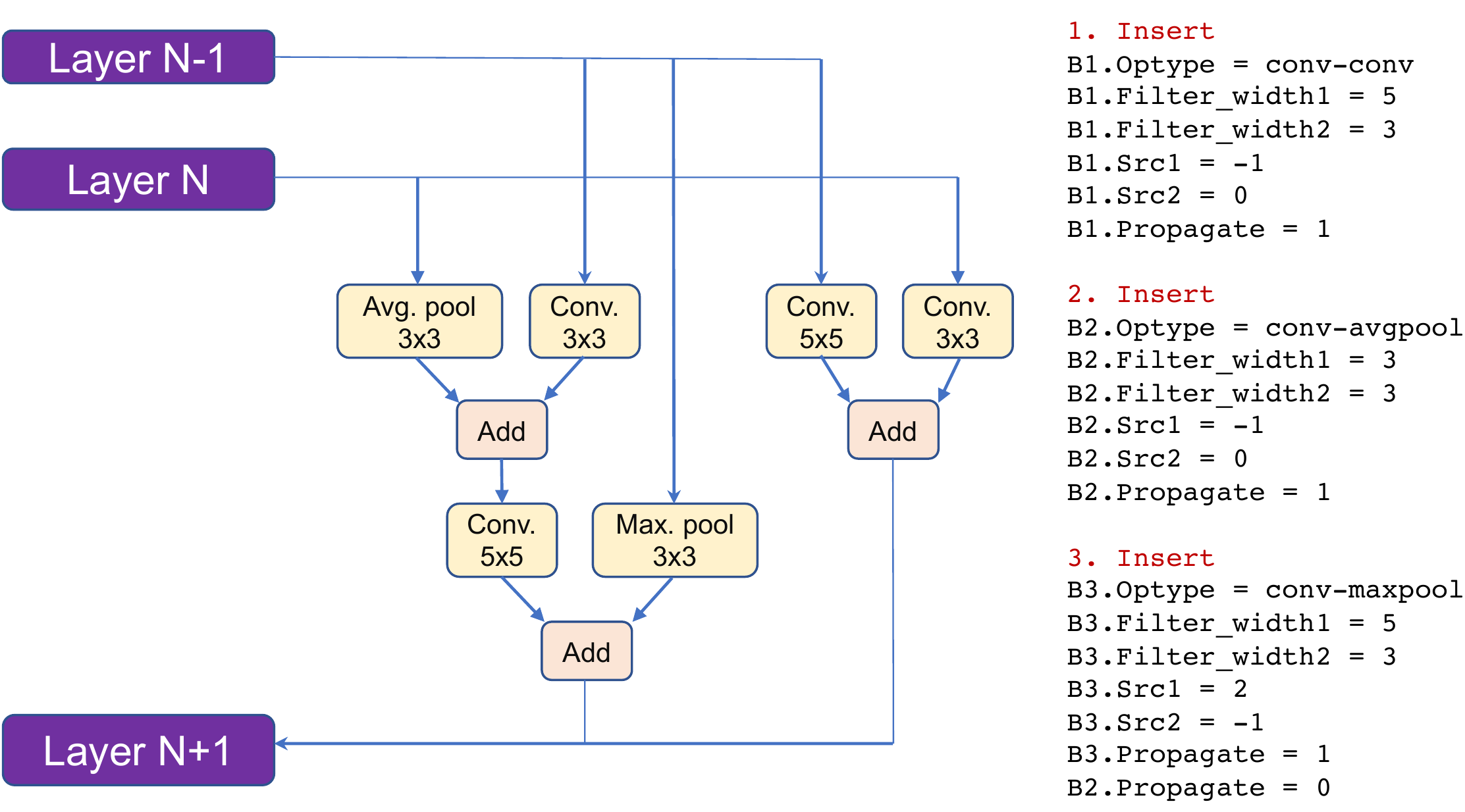}
        \caption{}
        \label{fig:module-insert}
    \end{subfigure}
    \caption{(a) An insert operation example for layer-by-layer search. (b) An insert operation example for module search. When the branch 3 is inserted, one of its source value is from branch 2. Once it is inserted, it cuts off the connection between branch 2 and the next layer, and sets ``propagate'' feature.}
    \vspace{-1em}
\end{figure}

\subsubsection{Layer-by-layer search}

Layer-by-layer search aims to find the optimal architecture with a search granularity of predefined layers. The neural network architecture is defined by stacking these layers, potentially with skip connections between them. For each feature, the LSTM in the policy network chooses the layer type and the corresponding hyperparameters (e.g. filter width). The location of the inserted layer is determined by ``Src1'', where the new layer gets its input data from. To support skip connection, the Insert Controller generates operation ``add'' that connects two layers (``Src1'' and ``Src2'') with either an addition or a concatenation operation. Figure~\ref{fig:two-ops} exemplifies two types of insert operations. The search space of remove action is only the list of "Src1" for all layers of the existing architecture. Therefore, it removes the selected layer from the network according to the selected "Src1". 

\subsubsection{Module search}

Module search aims to find an optimal small network module which can be repeatedly stacked to create the overall neural network. Module search enables searching for multi-branch networks while effectively limiting the search space. The insert action in module search no longer inserts a layer, but inserts a ``branch''. The LSTM in policy network chooses the types of the operation, and corresponding hyperparameters (filter width, pooling width, channel size, etc.) Each branch consists of two operations to be concatenated. ``Src1'' and ``Src2'' determine where these two operations get input values from. ``propagate'' determines whether the output of the branch gets passed to the next layer. Figure~\ref{fig:module-insert} exemplifies the insert operation for module search. 

\subsection{Policy Gradient with Multi-Objective Reward}

\begin{figure}[t]
\centering
\includegraphics[width=0.7\linewidth]{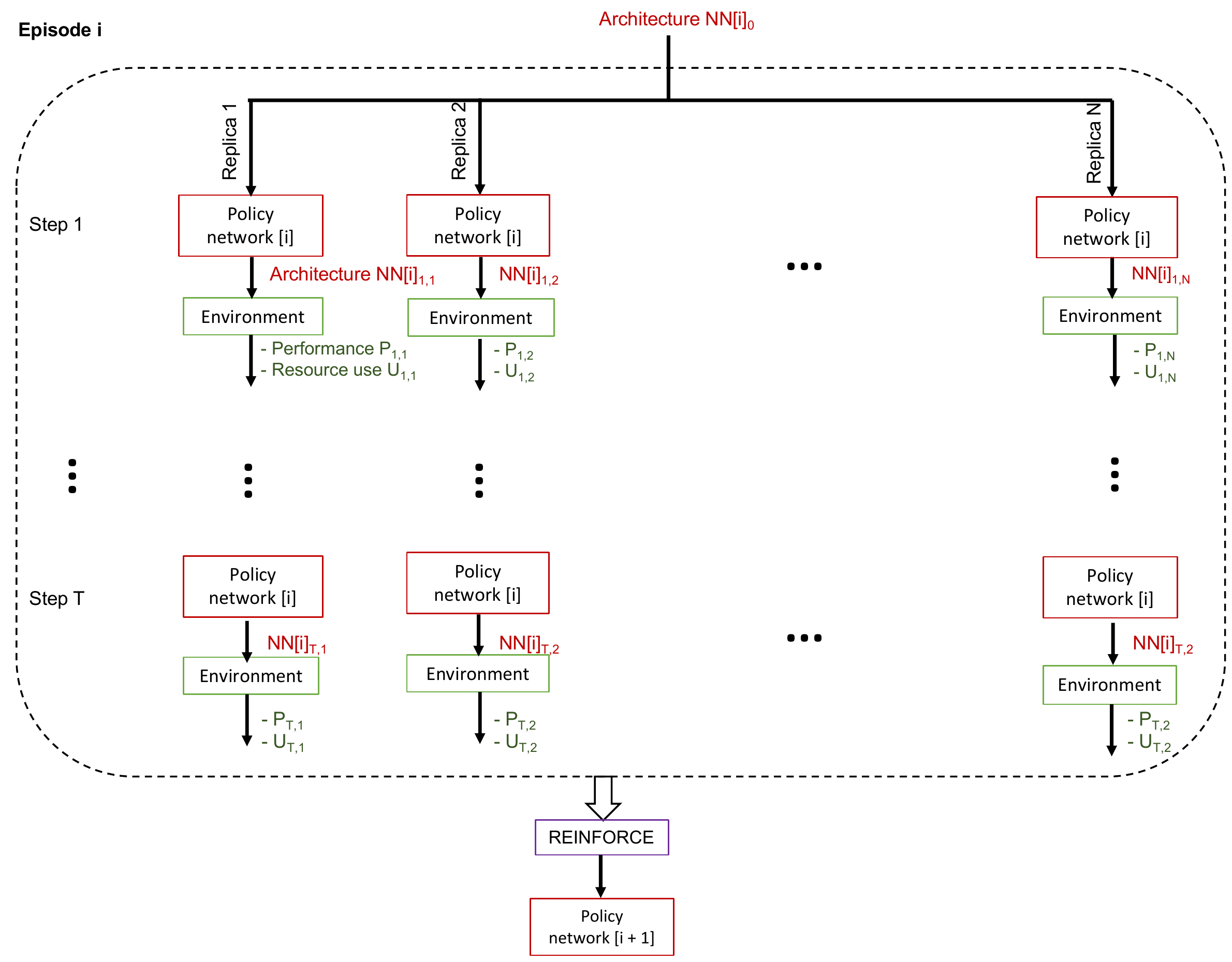}
\caption{Reinforce step for policy gradient. $N$ is the number of parallel policy networks to adapt a baseline architecture at episode of $i$.}
\vspace{-1em}
\label{fig:rl_algo}
\end{figure}

Fig. ~\ref{fig:rl_algo} depicts the reinforce step. The policy network generates a batch of actions $a_{t,n}$, which produce a series of child networks. We train child networks till convergence and use a combination of performance and resource use as an immediate reward, as given in Eq. \ref{reward}. Rewards of a full episode are accumulated to train the policy network using policy gradient~\citep{Williams.1992.PG}:
\begin{equation}
g = \frac{1}{N}\sum_{n=0}^{N-1}\sum_{t=0}^{T-1}\bigtriangledown_{\theta}\log\pi_{\theta}(a_{t,n}|s_{t,n})(\sum_{t^{'}=t}^{T-1}r_{t^{'},n}-b(s_{t})).
\end{equation}

To get a better estimate of whether the action is in the intended direction, returns should be compared to the expected. The difference $\sum_{t^{'}=t}^{T-1}r_{t^{'},n}-b(s_{t})$ is an approximate estimate of the efficacy of action $a_{t,n}$. To reduce the variance, we choose $b(s_{t})$ to estimate the expected sum of rewards:
\begin{equation}
\setstretch{1.5}
b(s) = \mathbb{E}[\sum_{t^{'}=t}^{T-1}r_{t^{'}} | s_{t}=s;a_{t:(T-1)\sim\pi_{\theta}}].
\end{equation}

In order to find neural architectures that meet multiple resource constraints, a reward based on the model performance should be penalized according to the extent of violating the constraints. Although a fixed hard penalty may be effective for some constraints, it would be challenging for the controller to learn from highly sparse rewards under tight resource constraints. Therefore, we use a soft continuous penalization method to enable finding architectures with high performance while still meeting all resource constraints. The reward $r$ for a specific architecture with performance $P$ and resource use $U$ (e.g. model size) when exposed to $M$ different resource constraints $C$ is determined by:

\vspace{-1em}
\begin{equation}
\label{reward}
    r = P\prod_{j=1}^{M}p^{V(U_{j}, C_{j})} \hspace{6pt} \textup{where} \hspace{6pt} V(U_{j}, C_{j}) = \left\{\begin{array}{lr}
        \max(0, U_{j}-C_{j})/C_{j} & \textup{Constraint:} \hspace{5pt} U_{j} < C_{j}\\
        \min(0, U_{j}-C_{j})/U_{j} & \textup{Constraint:} \hspace{5pt} U_{j} > C_{j}
        \end{array}\right.
\end{equation}

$V(U,C)$ is the violation function which determines the extent of violating a constraint depending on the type of the constraint. $p$ is the base penalty, which can be in range of 0 to 1. \footnote{In this work, we use $p$ = 0.9 for all architecture search experiments.}

\section{Experiments}

\subsection{Image Classification}

Image classification is one of the centerpiece problems of visual recognition applications, and it has been a competitive target for NAS given the successful results of highly-tuned neural network architectures. For image classification task, we consider the CIFAR-10 dataset. Standard image augmentation techniques, including random flipping, cropping, brightness and contrast adjustments, are applied. The performance is quantified in terms of the classification accuracy.

\subsubsection{Training Details}

The policy network is trained with the Adam optimizer with a learning rate of 0.0006. The weights of the controller are initialized uniformly between -0.1 and 0.1. At each step, 8 child models are constructed and trained for 150 epochs. We trained the child models with Nesterov momentum with a learning rate following the cosine schedule ($l_{max}=0.05, l_{min}=0.001, T_{0}=10, T_{mul}=2$)~\citep{SGDR}. For layer-by-layer search, we use an episode size of 10 and a batch size of 8. We progressively select top eight models from each episode as baseline models to the next episode. We train the best models for longer training time to get SOTA performance. For module search, we restrict the maximum number of branches to be five, as inserting more branches yields very long training time . We use an episode size of 5 and a batch size of 8. The baseline model with only one branch is always used as the baseline for all episodes. The search space is described in Appendix \ref{sec:searchspace}. An LSTM with 32 hidden units is used for network embedding, while larger LSTMs with 128 hidden units are used for Scale and Insert-Remove actions.

\subsubsection{Results}
\begin{figure}[t]
    \begin{subfigure}{0.5\linewidth}
        \centering
        \includegraphics[width=0.9\linewidth]{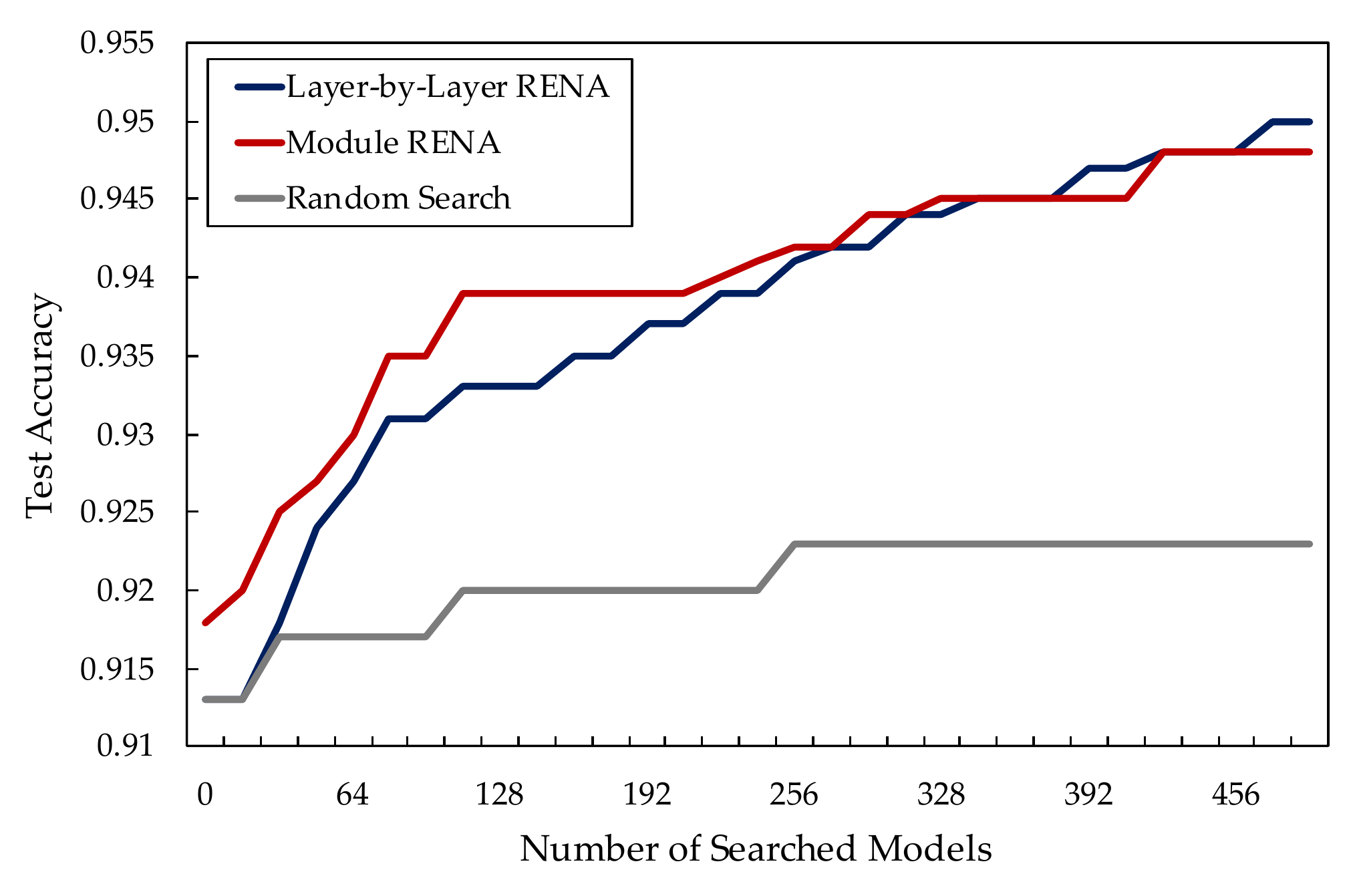}
        \caption{}
        \label{fig:na-accuracy}
    \end{subfigure}
    \begin{subfigure}{0.5\linewidth}
        \centering
        \includegraphics[width=0.95\linewidth]{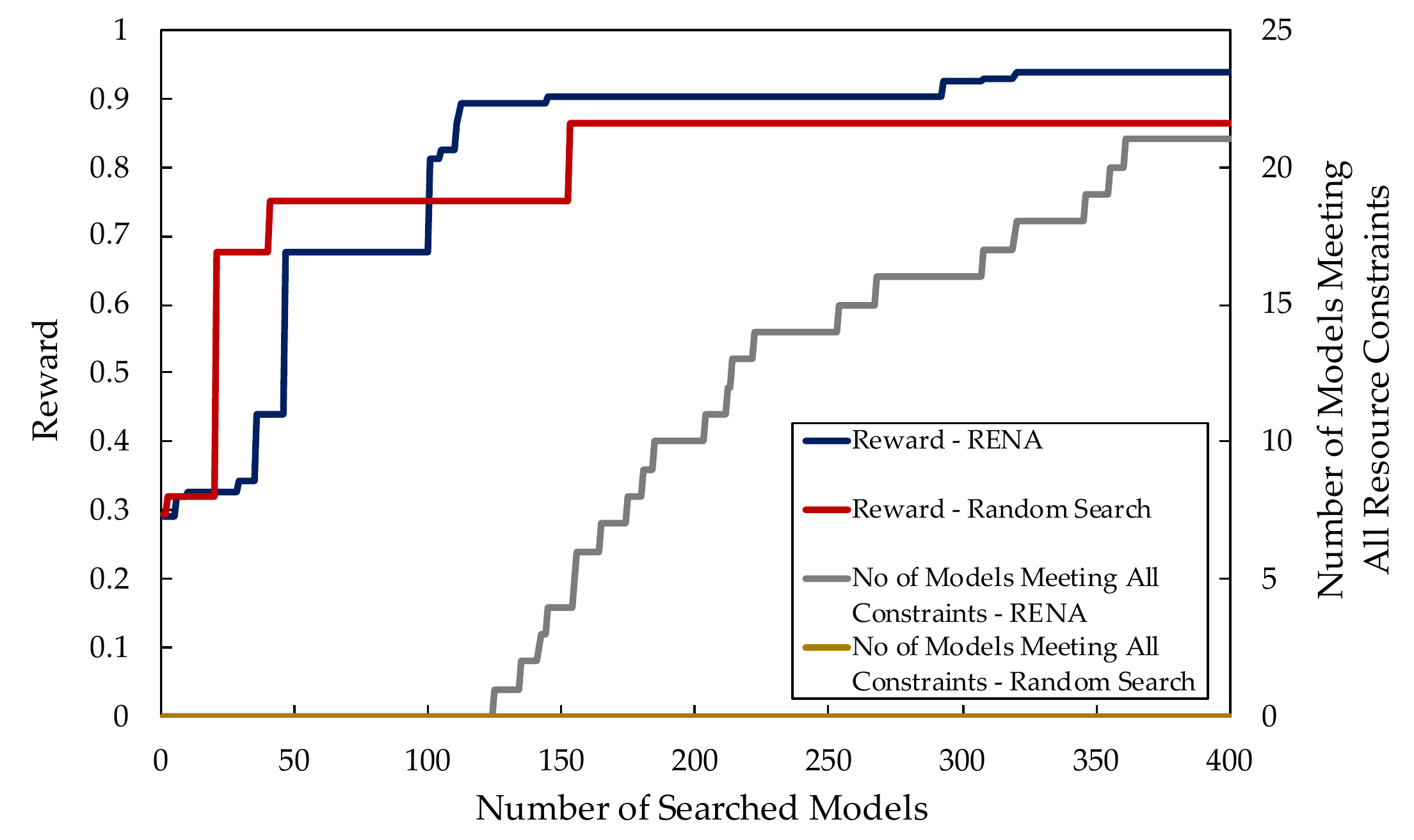}
        \caption{}
        \label{fig:RLvsRandom}
    \end{subfigure}
    \caption{(a) Best accuracy vs. total searched network architectures for CIFAR-10. (b) Comparison between RENA and Random Search for a resource-constrained architecture search (with model size < 0.1 M parameters and compute intensity > 10 FLOPs/byte) for Google Speech Commands Dataset.}
    \vspace{-1em}
\end{figure}

Fig. ~\ref{fig:na-accuracy} shows that RENA improves the test accuracy up to 95\% after 500 searched models, when started with a baseline model with a test accuracy of about 91\%. Both layer-by-layer search and module search significantly outperform random search. Table~\ref{tab:comparison} shows the comparison between RENA and SOTA models in accuracy, model size, and compute intensity. Popular SOTA models typically has high parameter counts and low compute intensity, compared to the best models found by RENA under resource constraints. More specifically, RENA is able to find model under 10M parameters with 3.48\% test error with 92 FLOPs/Byte compute intensity. With high compute intensity requirements, RENA finds models with large channel sizes and large filter widths. The network could even has large channel size for the first few convolution layers. With a tight constraint for both model size and compute intensity, RENA finds models with reasonably channel size that use a combination of depth-separable convolution and regular convolution. 


\begin{table}[ht]
  \centering
  \addtolength{\tabcolsep}{-4pt}
\caption{Comparison of RENA with the state-of-the-art models on CIFAR10.}
    \resizebox{5in}{!}{
    \begin{tabular}{|c|c|c|c|c|}
    \hline
    Model & Resource constraint & Parameters & Test error (\%) & \splitcell{Comp. intensity  \\ (FLOPs/byte) } \\ \hline\hline
    DenseNet (L=40, k=12) & - & 1.02 M & 5.24 & 4.1  \\ \hline
    DenseNet-BC (k=24) & - & 15.3 M & 3.62 & 7.1  \\\hline
    ResNeXt-29,8x64d & - & 34.4 M & 3.65 & 17.3 \\ \hline
    RENA: Layer-by-Layer Search& Model size < 10M & 7.7 M & 3.48 & 92\\ \hline
    RENA: Layer-by-Layer Search & Model size < 5M & 3.4 M & 3.87 & 42 \\ \hline
    RENA: Layer-by-Layer Search & Comp. intensity > 80 FLOPs/byte & 29 M & 2.95 & 107\\\hline
    RENA: Module Search & Model size < 3M & 2.2 M & 3.98 & 3.9 \\ \hline
    RENA: Module Search  & Model size < 5M & 4.0 M & 3.22 & 4.2 \\ \hline
    \end{tabular}
  }
  \label{tab:comparison}
\end{table}

\subsection{Keyword Spotting}

\begin{table}[!ht]
  \centering
  \addtolength{\tabcolsep}{-4pt}
\caption{Comparison of KWS models found by RENA and previous SOTA models. Conv2d (2-D convolution) and DS-Conv2d (2-D depth-separable convolution) are parametrized by the number of layers, channel size, kernel size in time and frequency, and stride in time and frequency, respectively. GRU is parametrized by the number of layers, number of hidden units, and the number of directions. FC (fully connected) is parametrized by number of layers and number of hidden units. AvgPool2d (2-D average pooling) is parametrized by pooling in time and frequency. }
    \resizebox{5in}{!}{
    \begin{tabular}{|c|c|c|c|c|c|c|c|c|}
    \hline
    Model & Resource constraints & Architecture &
    \splitcell{Parameters} & 
    \splitcell{Test \\ accuracy \\ (\%)} & 
    \splitcell{Compute \\ complexity  \\ (GFLOPs) } & 
    \splitcell{Compute \\ intensity  \\ (FLOPs/byte) } 
     \\ \hline\hline
    \splitcell{RENA: Layer-by-Layer Search} & - & \splitcell{DS-Conv2d (1,4,4,1,1,1) \\ GRU (1,64,1) \\ GRU (1,128,1) \\ Conv2d (1,12,16,2,4,4) \\ Conv2d (1,4,16,4,4,4) \\ Conv2d (1,64,16,4,4,4) \\ FC (1,32)} & 0.143M & 95.81 & 3.39 & 3.58  \\ \hline
    \splitcell{RENA: Layer-by-Layer Search} & Model size < 0.05 M & GRU (2,64,1) & 0.047M & 94.04 & 1.40 & 3.69  \\\hline
    \splitcell{DS-CNN \\ \citep{helloedge}} & - & \splitcell{Conv2d(1,64,10,4,2,2) \\ DS-Conv2d (4,64,3,3,1,1) \\ AvgPool2d} & 0.023M & 93.39 & 6.07 & 1.76  \\\hline
    \splitcell{RENA: Layer-by-Layer Search} & Model size < 0.1 M & 
    \splitcell{Conv2d (3,32,4,8,1,3) \\ AvgPool2d} & 0.067M & 94.82 & 6.53 & 8.11  \\ \hline
    \splitcell{RENA: Layer-by-Layer Search} & Comp. complexity < 1 GFLOPs & \splitcell{GRU (3,32,1) \\ FC (1,256)} & 0.425M & 93.16 & 0.89 & 2.45 \\\hline
    \splitcell{GRU \\ \citep{helloedge}} & - & GRU (1,154,1) & 0.093 M & 92.94 & 0.68 & 5.03 \\ \hline
    \splitcell{RENA: Layer-by-Layer Search} & Comp. complexity < 5 GFLOPs & 
    \splitcell{GRU (5,64,1) \\ FC (2,16)} & 0.171M & 95.02 & 3.30 & 6.38  \\ \hline
    \splitcell{RENA: Layer-by-Layer Search} & Comp. intensity > 10 FLOPs/byte & GRU (3,128,2) & 0.733M & 95.64 & 13.59 & 21.83  \\ \hline
    \splitcell{CRNN \\ \citep{helloedge}} & - & \splitcell{Conv2d (1,100,10,4,2,1) \\ GRU (2,136,1) \\ FC (1,188)} & 2.447M & 94.40 & 46.21 & 15.76  \\ \hline
    \splitcell{RENA: Layer-by-Layer Search} & Comp. intensity > 50 FLOPs/byte & 
    \splitcell{Conv2d (3,192,8,4,1,3) \\ AvgPool2d (8,1) \\ FC (2,16)} & 2.626M & 95.18 & 210.13 & 58.70 \\\hline
    \splitcell{RENA: Layer-by-Layer Search} & \splitcell{Model size < 0.1 M \\ Comp. intensity > 10 FLOPs/byte} & 
    \splitcell{Conv2d (2,32,20,2,1,2) \\ GRU (3,16,1) \\ GRU (2,12,1) \\ Conv2d (2,4,20,8,1,2)}& 0.074M & 93.65 & 12.57 & 10.29  \\\hline
    \splitcell{RENA: Layer-by-Layer Search} & \splitcell{Model size < 0.1 M \\ Comp. complexity < 1 GFLOPs} & GRU (2,32,2) & 0.035M & 93.07 & 1.00 & 2.77  \\\hline
    \end{tabular}
  }
  \label{tab:comparison-kws}
\end{table}

Keyword spotting (KWS) systems aim to detect a particular keyword from a continuous stream of audio. They are commonly used in conversational human-machine interfaces, such as in smart home systems or virtual assistants. A high detection accuracy and a low latency is critical to enable satisfactory user experience. In addition, KWS systems are typically deployed on a wide range of devices with different resource constraints. Therefore, an optimal resource-constrained neural architecture design is very crucial.

For KWS task, we use the Google speech commands dataset \citep{speechcommands}. Similar to \citep{helloedge}, we consider KWS problem with 12 classes. The dataset split is also similar to \citep{helloedge} that training, validation and test sets have the ratio of 80:10:10 while making sure that the audio clips from the same person stays in the same set. The performance is quantified in terms of the classification accuracy. Further details are given in Appendix \ref{preprocesskws}.

\subsubsection{Training Details}

For KWS architecture search, only layer-by-layer search is considered, while always starting from a small baseline architecture, i.e. a single fully connected layer with 12 hidden units, which yields a test accuracy of 65\%. The policy network is trained with the Adam optimization algorithm with a learning rate of 0.0006. An episode size of 5 and a batch size of 10 is used for all experiments, i.e. 10 child models are trained concurrently. The search space is given in Appendix \ref{sec:searchspace}. Each model is evaluated after training and and an action is selected according to the current policy in order to transform the network. At the end of each episode, the policy is updated and the best 10 child models are used as the baseline for the new episode. The weights of the controller are initialized uniformly between -0.1 and 0.1. The size of LSTMs for network embedding and the controllers are similar to those of the image classification task.

\subsubsection{Results}

Fig. ~\ref{fig:RLvsRandom} compares the effectiveness of RENA versus Random Search\footnote{The Random Search generates random actions (i.e. insert, remove, and scale) and also selects the hyperparameters of each layer randomly with uniform probability.} in finding a resource-constrained architecture (size < 0.1 M and compute intensity > 10) in terms of reward and number of models meeting both constraints. As illustrated in the figure, RENA learns to generate models that meet both constraints after about 120 searched models whereas Random Search is not able to generate any models meeting both constraints within 400 searched models. RENA attempts to maximize the model performance in this domain and finally finds an architecture with 93.65 \% test accuracy that meets both resource constraints. Random Search can barely find a model that violate the constraints by a small margin (model size = 0.13 M and compute intensity = 10.69 FLOPs/byte). 

Table~\ref{tab:comparison-kws} presents the search results for KWS, as well as the optimal architectures. Without any resource constraints, the state-of-the-art accuracy, 95.81\%, can be obtained using an architecture composed of depth-separable convolutions (that apply significant downsampling), followed by GRUs and multiple 2-D convolutions. When aggressive resource constraints are imposed, we observe that RENA can find architectures that outperform hand-optimized architectures in the literature. A tight model size constraint results in an optimal architecture composed of GRUs with small hidden units. Similarly, tight constraints on computational complexity also favor for GRUs with small hidden units. When compute intensity is considered, an efficient architecture is achieved by enabling most of the computation on 2-D convolutions with large channel size. Lastly, we consider joint constraints, and we observe that very competitive accuracy results can be obtained even in the regime of a small feasible architecture space. For example, RENA finds models under 0.1M parameters with high compute intensity (>10 FLOPs/Byte) with 93.65\% test accuracy. We observe the benefits of high compute intensity and low computational complexity for low inference latency, as expected.  

\section{Conclusions}
We propose RENA, an resource-constrained NAS with network embedding, achieved using RL. Policy network is designed to process the network embedding by predefined actions to create new network configurations. Our framework can achieve sample-efficient search - RENA achieves >95\% accuracy for CIFAR10 within 500 total searched models. Besides, we demonstrate a framework to integrate resource constraints in automated NAS. Constraints are imposed by modifying the reward function to penalize cases when the generated models violate the constraints. We demonstrate that our model can achieve very competitive results for image recognition (on CIFAR10 Dataset) and keyword spotting (on Google Speech Commands Dataset) even with tight constraints.

\newpage

\bibliographystyle{abbrvnat}
\bibliography{bibliography}

\newpage
\pagebreak
\appendix
\appendixpage

\section{Complexity Modeling}
\label{complexity_modeling}

Complexity of mathematical operations is represented by the total number of algorithmic FLOPs without considering hardware-specific logic-level implementations. Such a complexity metric also has limitations of representing some major sources of power consumption, such as loading and storing data. 

We count all point-wise operations (including nonlinearities) as 1 FLOP, which is motivated with the trend of implementing most mathematical operations as a single instruction. We ignore the complexities of register memory-move operations. We assume that a matrix-matrix multiply, between $W$, an $m \times n$ matrix and $X$, an $n \times p$ matrix takes $2mnp$ FLOPs. Similar expression is generalized for multi-dimensional tensors, that are used in convolutional layers. For real-valued fast Fourier transform (FFT), we assume the complexity of $2.5 N log_2(N)$ FLOPs for a vector of length $N$ \footnote{http://www.fftw.org/speed/method.html}. For most operations used in this paper, Tensorflow profiling tool \footnote{https://github.com/tensorflow/tensorflow/blob/master/tensorflow/core/profiler/README.md} includes FLOP counts, which we directly used. 

\section{Training details for KWS models}
\label{preprocesskws}
The raw time-domain input audio samples have a duration of 1 second, sampled at a rate of 16 kHz. Speech features are extracted using 40 Mel-frequency cepstral coefficients (MFCC) with a hop length of 20 ms and a window length of 40 ms, yielding 2-D spectrograms with dimensions of $49\times40$. Random time-jittering of 100 ms is applied for augmentation. In addition, 80 percent of training and test samples are augmented by applying additive noise with a signal-to-noise ratio (SNR) in range of [10,20] dB, sampled from the background noise data in the dataset. 

The ADAM optimization algorithm is used for training each KWS model, with a batch size of 128 and an initial learning rate of 0.001. The learning is dropped by 0.2 every 10000 training iterations. Due to the small scale of the problem, a cross entropy (CE) loss function is used for training. 

\section{Search Space}
\label{sec:searchspace}
Table~\ref{tab:insert_space_image}, Table~\ref{tab:insert_space_kws}, and Table~\ref{tab:insert_space_c} demontrates the search space for our image recognition and KWS. 

\begin{table}[!htp]
  \centering
  \addtolength{\tabcolsep}{-5pt}
\caption{Search space of scale and insert actions in layer-by-layer search for image classification.}
    \resizebox{4in}{!}{

    \begin{tabular}{|c|c|}
    \hline
    Feature &  Search space \\ \hline\hline
    Layer type & [conv2d, dep-sep-conv2d, MaxPool2d, add] \\\hline
      Filter width & [3,5,7] \\ \hline
      Pooling width & [2,3] \\ \hline
      Channel size & [16,32,64,96,128,256] \\\hline
      Nonlinear activation & ["relu","crelu","elu","selu","swish"] \\\hline
      Src1 Layer & [i for i in range(MAX\_LAYERS)] \\\hline
      Src2 Layer & [i for i in range(MAX\_LAYERS)] \\\hline
      
    \end{tabular}
  }
  \label{tab:insert_space_image}
\end{table}

\begin{table}[!tp]
  \centering
  \addtolength{\tabcolsep}{-5pt}
\caption{Search space of scale and insert actions in layer-by-layer search for keyword spotting.}
    \resizebox{4in}{!}{

    \begin{tabular}{|c|c|}
    \hline
    Feature &  Search space \\ \hline\hline
    Layer type & [conv2d, dep-sep-conv2d, dilated-conv2d, \\
               & GRU, AvgPool2d, FC] \\\hline
      Number of Layers & [1, 2, 3, 4, 5] \\ \hline
      Kernel size in time & [1, 4, 8, 16, 20] \\ \hline
      Kernel size in frequency & [1, 2, 4, 8, 10] \\ \hline
      Channel size (or hidden units) & [4, 12, 16, 32, 64, 128, 192, 256] \\\hline
      Stride in time & [1, 2, 4, 8, 10] \\\hline
      Stride in frequency (or dilation rate) & [1, 2, 3, 4, 5] \\\hline
      Number of GRU directions & [1, 2] \\\hline
      Dropout rate & [0.8, 0.9, 1.0] \\\hline
      Src1 Layer & [i for i in range(MAX\_LAYERS)] \\\hline
      Src2 Layer & [i for i in range(MAX\_LAYERS)] \\\hline
      
    \end{tabular}
  }
  \label{tab:insert_space_kws}
\end{table}
\begin{table}[!tp]
  \centering
  \addtolength{\tabcolsep}{-5pt}
\caption{Search space for scale and insert actions in module search for image classification.}
    \resizebox{4in}{!}{

    \begin{tabular}{|c|c|}
    \hline
    Feature &  Search space \\ \hline\hline
    Branch type & [conv-conv, conv-maxpool, conv-avgpool, \\
                & conv-none, maxpool-none, avgpool-none,$1\times7$-$7\times1$-none] \\\hline
      Filter width & [3,5,7] \\ \hline
      Pooling width & [2,3] \\ \hline
      Channel size & [8,12,16,24,32] \\\hline
      Src1 Layer & [i for i in range(MAX\_BRANCHES+1)] \\\hline
      Src2 Layer & [i for i in range(MAX\_BRANCHES+1)] \\\hline
      Propagate & [0,1] \\\hline
      
    \end{tabular}
  }
  \label{tab:insert_space_c}
\end{table}

\end{document}